# Micro-Doppler Based Human-Robot Classification Using Ensemble and Deep Learning Approaches

Sherif Abdulatif*, Qian Wei*, Fady Aziz, Bernhard Kleiner, Urs Schneider
Department of Biomechatronic Systems, Fraunhofer Institute for Manufacturing Engineering and Automation IPA
Email: {sherif.abdulatif, qian.wei, fady.aziz, bernhard.kleiner, urs.schneider}@ipa.fraunhofer.de
* These authors contributed to this work equally.

*Abstract*—Radar sensors can be used for analyzing the induced frequency shifts due to micro-motions in both range and velocity dimensions identified as micro-Doppler ($\mu$-D) and micro-Range ($\mu$-R), respectively. Different moving targets will have unique $\mu$-D and $\mu$-R signatures that can be used for target classification. Such classification can be used in numerous fields, such as gait recognition, safety and surveillance. In this paper, a 25 GHz FMCW Single-Input Single-Output (SISO) radar is used in industrial safety for real-time human-robot identification. Due to the real-time constraint, joint Range-Doppler (R-D) maps are directly analyzed for our classification problem. Furthermore, a comparison between the conventional classical learning approaches with handcrafted extracted features, ensemble classifiers and deep learning approaches is presented. For ensemble classifiers, restructured range and velocity profiles are passed directly to ensemble trees, such as gradient boosting and random forest without feature extraction. Finally, a Deep Convolutional Neural Network (DCNN) is used and raw R-D images are directly fed into the constructed network. DCNN shows a superior performance of 99% accuracy in identifying humans from robots on a single R-D map.

## I. INTRODUCTION

Identifying a moving target and classifying its motions are grasping great interest in many fields. Numerous safety and surveillance applications now require systems that can be used for human motion identification, which are known as biometric systems. Due to recent development in the field of industry, robots are now widely used for assembly line automation. Therefore, the need of a safe Human Robot Interaction (HRI) environment is highly needed to avoid potential threats due to robots powerful movements [1]. In order to provide a safe HRI in such unstable environments, a reliable, real-time and accurate human detection system is required.

Camera-based systems are used for human detection based on feet and head recognition of the objects in the scene [2]. In [3], infrared cameras are used to enhance night vision human detection for surveillance applications. However, all of the vision-based sensors suffer limitations in different lighting and weather conditions. In industrial safety applications, the standardized sensor for safe human detection is LIDAR (Light Detection and Ranging), where reflections from human legs at knee level are processed to identify presence of a human in the scanned area [4]. LIDAR sensors, though have many limitations in detecting reflections from dark surfaces and problems with outdoor harsh environmental conditions.

Unlike other vision based sensors, radar systems can still detect targets behind obstacles or hard surfaces and can work in harsh outdoor circumstances [5]. Due to developments in radar technology, especially in the micro-Doppler ($\mu$-D) field, radar can be used nowadays to extract target's bulk motion (Range-Velocity) in addition to micro-motions as limbs swings or even very fine vibrations due to vital signs [6]. Recently, $\mu$-D effect in radar has been extensively addressed in different applications related to biometric systems, such as gait recognition and limbs decomposition [7]. The author in [8] focused on distinguishing human target from different objects, such as animals based on their corresponding $\mu$-D signatures. Others focused on using radar for safe pedestrian recognition in automotive industry to apply an emergency brake to the car when a pedestrian is detected at a close distance [9]. Most of the previously mentioned papers applied human detection based only on velocity analysis, where CW radars were used to extract $\mu$-D signatures of the whole human gait cycle i.e, more than 1 second duration. Signatures are used with either conventional machine learning or Deep Convolutional Neural Networks (DCNN) approaches on the collected dataset to detect possible human presence.

However, these techniques did not address the real-time classification aspect. Moreover, the size of the training dataset proposed for designing a classifier is in the order of hundreds. Such amount is not enough for designing robust classifiers, especially in DCNN, as it is proven that for the training of a DCNN, more training data can increase the model performance [10]. In this paper, the human-robot classification problem is addressed on collected FMCW radar Range-Doppler (R-D) maps that can be computed in much shorter time intervals (tenth of a second) [11]. Accordingly, real-time constraint can be fulfilled and the amount of data used to train deep models increased from hundreds to thousands of datasets. Finally, a comparison between different learning approaches is presented.

The paper is organized as follows: Section II introduces the dataset preparation and R-D map computations. Section III shows the use of conventional learning approaches on hand crafted features. A novel data-driven approach based on ensemble tree classifiers is presented in Section IV. In Section V, a DCNN is designed and applied on R-D maps as images. Finally, all approaches are compared with suggested future work in Section VI.

## II. DATASET PREPARATION

As mentioned above, R-D maps are used to build the dataset for all presented learning approaches. A procedure for R-D mapping was proposed in [12]. As illustrated in Fig. 1, the procedure applies Fast Fourier Transform (FFT) to each measurement with a specified FFT length ($N_{FFT}$) on both dimensions of the data matrix. Each column stands for one chirp with $N_s$ samples, while one measurement consists of $N_p$ chirps. Accordingly, each R-D map can still contain abundant information about the $\mu$-D and the $\mu$-R of the moving object, as long as the range and velocity resolutions of the radar are sufficiently high. A 512 FFT is used to achieve acceptable resolution on both dimensions. After FFT, the range information is estimated within each chirp, while the Doppler data is estimated across all chirps in one sample. In this way, one R-D map can be acquired from each data matrix. Each element in the data matrix specifies the back-scattered power at these particular range and velocity. These maps can be visualized as a (512 × 512) heat map, where Doppler is represented on the horizontal axis, ranges on the vertical axis and the color RGB values from 0 to 255 represent the back-scattered power.

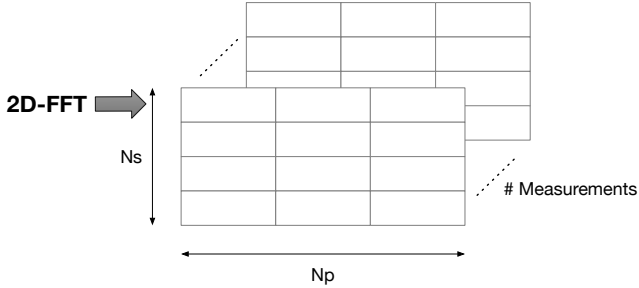

Fig. 1: R-D mapping procedure.

*A. Radar Parametrization*

The FMCW radar used for data collection operates at a carrier frequency $f_c = 25\,\text{GHz}$ at a maximum operating bandwidth of $B = 2\,\text{GHz}$ with a chirp sweeping time of $T_p = 0.5\,\text{ms}$. Based on the given bandwidth and Eq. 1, a reasonable range resolution for our application is derived as $R_{res} = 7.5\,\text{cm}$, where $c$ is the speed of light. Since the proposed human robot detection is required on a room level, the maximum measurement range is limited to $R_{max} = 5\,\text{m}$. Accordingly, the number of samples per chirp ( see Eq. 2) can be calculated as $N_s \approx 67$ samples.

$$R_{res} = \frac{c}{2B} \quad (1)$$

$$N_s = \frac{2BR_{max}}{c} \quad (2)$$

In [7], infrared motion captured data was collected on different walking human subjects and it was proved that the feet has the maximum swinging velocity component that can reach up to 4.5 m/s. Based on Eq. 3, a maximum velocity $v_{max} = 6$ m/s can be achieved with the given carrier frequency and chirp duration. This shows that our attained maximum velocity can safely cover human walking test subjects. For the requirement of human robot classification, a velocity resolution of $v_{res} = 0.1$ m/s is proposed. According to Eq. 4, the number of chirps per measurement to the next power of 2 is $N_p = 128$ chirps. Using the given chirp duration ($T_p$), one R-D map of $N_s \times N_p$ size will theoretically take a measurement duration of 64 ms. After taking processing delays into consideration, a final measurement duration of one R-D map is still lower than 0.1 s. This duration is far better for a real-time classification constraint. Based on the proposed parameters, both range and velocity resolution are high enough to induce R-D maps that contain sufficient information about the $\mu$-D and the $\mu$-R of a complex human target. All presented radar design equations are derived in [12].

$$v_{max} = \frac{c}{4f_cT_p} \quad (3)$$

$$N_p = \frac{c}{2f_cT_pv_{res}} \quad (4)$$

*B. Dataset Build-up*

To collect the data, measurements have been carried out on 10 test walking human subjects including different heights and genders. There exist uncountable types of machines under the name "robot". Since industry is addressed for the proposed task, two industrial robots are included as subjects for data collection and testing. The first robot is a moving mobile robot assistant (Care-O-bot) developed by the robotics department in Fraunhofer IPA [13]. The second robot is a 6-axis robotic arm developed by Stäubli [14]. The subjects were moving in random aspect angles in the radar detectable area during each experiment. However, the case of exact lateral motion was not included to avoid the extreme radial velocity fading effects mentioned in [9].

During the data collection, each data sample is labeled with the current target class. This labeling is required for supervised learning to allow models to know the "standard solution", and thus learn the right model parameters. The collected data together with the labels is divided into two subsets. The first subset will be used for construction of the model. This subset is then divided into two parts, one part called "training set" is used to learn the model parameters, while the other part named "validation set" is used to tune the model hyper-parameters, such as the number of hidden layers and neurons of each layer in a neural network. The second subset (test set) is employed to assess the generalization performance of the final model by comparing the model predictions with the true class labels.

The hold-out method is used to create the test set. For a hold-out split, the complete dataset $D$ is divided into two disjoint subsets. One is used as the training set $S$, the other is employed as test set $T$. Mathematically, it can be expressed as: $D = S \cup T, S \cap T = \varnothing$. The models are trained on $S$ and tested on $T$ to gain a glimpse about the generalization performance. Since contiguous samples from one measurement can be very similar due to the relatively high sampling frequency, a test on randomly chosen samples from all measurements cannot reflect the true generalization performance of the model. Consequently, separate experiments of each object type is chosen to build the test set, such that it takes up to 20% of the entire data of this type. By this means, the whole test set contains around 20% of all collected data. Finally, a human-robot dataset with 7740 training samples and 1892 test samples is obtained, in both of which the amount of human samples and robot samples are comparable.

*C. R-D Interpretations*

The human walking motion is described as successive periodic cycles in which two phases can fully describe a gait cycle [6]. The first phase (swinging) in which there is only one swinging foot and the other is touching the ground. Within this phase, the human appears on the R-D map as a broad distribution in both R-D axes as shown in Fig. 2a. This distribution represents a variety of velocities due to the bulk moving body parts (torso and head), in addition to the swinging effect of different body limbs (arms, legs and feet). In the second phase (stance), no limbs swinging and only bulk motion is observed. This phase is the dominant phase occupying 60% of the gait cycle. By comparing robot motion and human stance phase shown in Fig. 2b and Fig. 2c, respectively. It is clear that the stance phase represents the main challenge in our human-robot differentiation task, since both R-D maps are very similar in the narrow horizontal distribution.

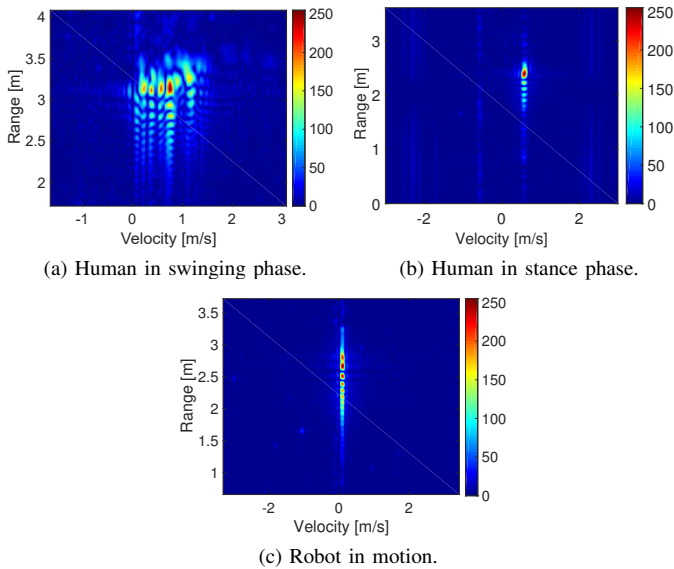

Fig. 2: Comparison between R-D maps of human and robot.

## III. CONVENTIONAL LEARNING ON HANDCRAFTED FEATURES

After obtaining the dataset as described in the previous section, we tested several conventional machine learning methods with hand-crafted features extracted from the R-D maps. Before the feature extraction, the multi-Otsu method is used on R-D maps as an unsupervised image thresholding to extract the R-D data corresponding to the target from the background [15]. By leveraging it, the original continuous RGB values in each R-D map are quantized into 10 discrete levels. The lowest 5 levels from the 10 are neglected since they can be considered as noise. Fig. 3 illustrates the effect of applying multi-Otsu method to the R-D map of a human swinging phase shown in Fig. 2a.

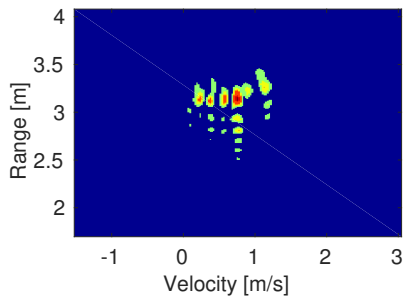

Fig. 3: The R-D map of human after multi-Otsu thresholding.

Subsequently, features are extracted from each R-D map to represent the "Human/Robot" differences discussed previously. The features are also chosen, such that they do not reflect an exact velocity or range to avoid overfitting. Thus, the extracted features only considers the distribution over R-D maps. The distribution can be represented as two features for the range and velocity profiles, which are computed as the differences between the maximum and the minimum detected values in both R-D dimensions. Moreover, features as the variance in velocities $\sigma_v^2$ and in ranges $\sigma_R^2$ can be seen as the polynomial features of the standard deviation in both velocities $\sigma_v$ and ranges $\sigma_R$ with a degree of 2 [16]. Furthermore, the covariance between range and Doppler values is also considered as a feature. Finally, we have 7 features used with different conventional machine learning techniques.

After the feature extraction, several classical machine learning methods have been implemented, trained and evaluated with the feature data. The employed methods are (a) Decision Tree, (b) Logistic Regression, (c) Support Vector Machine (SVM) and (d) K-Nearest Neighbors (K-NN). However, depending on features from one single R-D map, the performance of all methods is less than satisfactory. This conduced to the use of feature vector sequence accumulated from several successive R-D maps (a so-called sample buffer). Each feature vector sequence is concatenated by feature vectors extracted from all successive samples in a buffer. Accordingly, the number of features used will increase from 7 (in the case of a buffer of size 1) to 70 (in the case of a buffer of size 10). As shown in Fig. 4, the classification accuracies of all the methods increase as the buffer size increases. The best test accuracy is reached using SVM as 95.3% at buffer size 10. However, such large buffer size induces a latency of more than 1 second, which leads to problems in safety critical real-time applications.

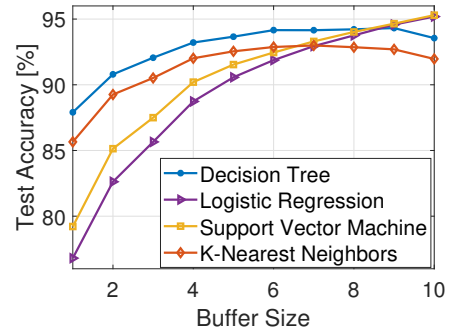

Fig. 4: Accuracy curves of conventional learning methods.

## IV. ENSEMBLE TREES WITH RESTRUCTURED R-D DATA

There are two main drawbacks of using conventional machine learning methods with hand-crafted features. On one hand, manual construction of features from raw data is time-consuming and requires sufficient domain knowledge. On the other hand, as described in the previous section, sufficient performance can only be obtained at the cost of a large buffer size, which directly correlates with the inference latency.

### A. Ensemble Learning

Generally, the predictive power acquired by combining a bunch of models is better than only using one single model.

An Instance of this idea is "ensemble learning", a family of machine learning methods which perform the learning task by constructing a group of individual learners and combine their outputs together as the final output.

In machine learning, one must always be faced with the bias-variance trade-off. Bias and variance are two types of error of a predictive model. A simple model is prone to underfitting of the training data; thus, having high bias, but low variance. Conversely, complex models tend to overfit the training set and thus having low bias, but large variance. From this dilemma, two opposite procedures can be proposed for decreasing prediction error: reducing the variance of complex models and reducing the bias of simple models.

Krogh and Vedelsby proved in [17] that the error of an entire ensemble $\widehat{E}$ can be determined by:

$$\widehat{E} = \overline{E} - D \quad (5)$$

where $\overline{E}$ represents the average error of all individual learners, while $D$ evaluates the degree of diversity of individual learners. This indicates that, in order to reduce the predictive error of an ensemble model, the individual learners should be diverse.

Two common types of ensemble learning are bagging and boosting. To realize the diversity of individual learners, both bagging and boosting leverage varying training sets, on which individual learners are trained. The difference lies in how the varying training sets are obtained.

**Bagging** is the abbreviation of *bootstrap aggregating*, which decreases the prediction error by reducing the variance of complex individual learners. In bagging, varying training sets are built by randomly sampling from the whole dataset with replacement (bootstrap sample). After building a predefined $K$ training sets, $K$ individual learners will be trained on these $K$ training sets. This means that, the individual learners can be generated in parallel; hence, there is no strong dependency between them. The hypothesis of the entire ensemble can be acquired by unweighted averaging of the hypotheses of all $K$ individual learners. Therefore, the estimated bias remains unchanged, while the estimated variance decreases by a factor of $K - 1$ [16].

**Boosting** improves the prediction performance by reducing the bias of weak individual learners. It constructs diverse training sets by iteratively assigning weights to data samples. The weight, with which each data sample is attached depends on how well this data sample can be predicted by the current ensemble. By doing so, the training data distribution is modified, and thus resulting in more attention to the data portion which is not well predicted so far.

Weak learners are simple models which can learn the training data with an accuracy not much higher than 50% (with a high bias and a low variance). Schapire proved in [18] that a group of weak learners could be combined into a strong ensemble achieving arbitrarily high training accuracy. Boosting employs this idea and constructs a set of weak learners sequentially. Each individual weak learner is induced with the current weighted training set obtained in the manner described previously. After generating the predefined number of individual $K$ learners, the ensemble hypothesis is obtained by weighted vote of predictions made by all weak learners. By combining weak learners that focus more on currently mis-predicted samples, both bias and variance of an ensemble will gradually decrease.

*B. Restructure of R-D Map*

Before feeding the classifiers, the obtained R-D map should be first restructured. Firstly, R-D map is of a two-dimensional structure $512 \times 512$, which must be "flattened" to a one-dimensional feature vector before feeding into the ensemble classifiers. To convert R-D maps into feature vectors, elements of each R-D map are averaged along both dimensions. This results into a two 512-dimensional vectors obtained; one representing the Doppler profile (row vector) and the other one representing the range profiles (column vector).

Secondly, to guarantee a rational classification based on target motion dynamics, instead of the absolute values, such as velocity or range. The information regarding absolute measurement values should be removed from the data, such that only the R-D distributions would be used. We propose a method to eliminate such information containing concrete "target motion parameters" as follows: in both Doppler and range profile vectors, the elements corresponding to high power areas (distribution) are shifted to the middle of each vector. Since the positions of these elements in both Doppler profile and range profile correlate to the absolute velocity and range of the target, respectively. By shifting the large-valued elements in both vectors to the middle, clues to the velocity and range of the target are eliminated. This algorithm works by normalizing the power values of both velocity and range profiles to the sum of all of their elements, respectively, to get a weights vector for each. Then, a weighted average is applied to both velocity and range indices based on their corresponding weights, to get an average value close to the high power area in each profile. Accordingly, the profiles are shifted from the computed indices to the middle of the vectors. To reduce the complexity, 128 elements are removed from both ends of the shifted vectors to get 256 elements per profile. After the dimension reduction, one 512-dimensional feature vector is built by concatenating both restructured Doppler and range profiles.

*C. Performance*

To study the feasibility of applying ensemble learning to restructured R-D maps, random forest and gradient boosting are tested. They can be regarded as outstanding representatives of bagging and boosting, respectively. As shown in Table I, the random forest achieves a worse performance compared to the gradient boosting. This can be explained by two reasons: (a) According to [19], boosted trees perform better than the random forest for a low dimensionality problem with a data dimension up to 4000. (b) Gradient boosting can achieve better results than random forest for a binary classification problem [20]. Thus, gradient boosting is considered as a better choice for further investigations and comparisons.

TABLE I: Accuracy of random forest and gradient boosting.

|  | Training Accuracy | Test Accuracy |
| --- | --- | --- |
| Random Forest | 99.88% | 93.40% |
| Gradient Boosting | 100.00% | 97.85% |

## V. CNN WITH R-D MAPS

The CNNs proved to be extremely effective in image classification, and since R-D maps are essentially images as well. Therefore, the application of CNNs is reasonable in this scenario. As previously illustrated in Fig. 2, a typical R-D map of human has a broad horizontal distribution which represents a variety of velocities of different body parts. Through comparison, one can see that the R-D map of a robot will only have a narrow horizontal distribution due to its rigid body motion. For human eyes, the difference between both patterns is already distinguishable. Accordingly, a CNN is able to differentiate them as well.

For the proposed CNN model, an input image size of $(200 \times 200)$ is used. This provides a trade-off between performance and processing time. The grayscale color mode is employed due to the following reason: in R-D maps, the color represents the back-scattered power which correlates to the target position relative to the radar and can be affected by metallic parts on targets (e.g., wearable metal articles such as watches or rings on human targets). In our approach, all of this information should not be considered. Compared to the grayscale, the RGB color mode is more sensitive to the noise resulting from unwanted objects and clutters. Furthermore, by using one-channel grayscale images (shown in Fig. 5) as input, the computational complexity of both training and prediction is reduced.

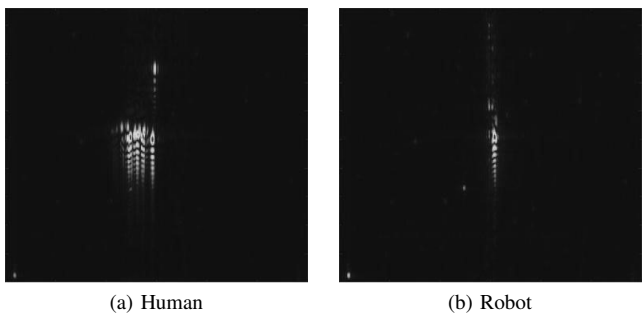

(a) Human      (b) Robot

Fig. 5: Grayscale R-D maps of human and robot fed to CNN.

### A. Network Architecture and Training

The network architecture used in our approach is inspired by Lecun's "LeNet-5" [21]. As illustrated in Fig. 6, it contains a stack of 6 convolutional layers with Rectified Linear Unit (ReLU) activation function. Each convolutional layer consists of 16 convolutional kernels with a size of $3 \times 3$ and is followed by a max-pooling layer. Furthermore, there is a fully-connected layer consisting of 16 neurons after the convolutional layer stack. The output layer at the end has one neuron with sigmoid activation which is fully connected to the 16 neurons of the previous layer.

The choice of the optimizer has an immediate effect on the result of the training, as well as, the required time. For the training of our proposed CNN, the modern adaptive optimizer Adam proposed in [22] is employed. The Adam optimization algorithm is currently one of the most popular algorithms for training various types of Deep Neural Networks (DNN), such as CNNs in computer vision applications and Recurrent Neural Networks (RNNs) for natural language processing. It enhances the classical stochastic gradient descent algorithm by enabling the computation of individual adaptive learning rates for different parameters. In this manner, the Adam optimizer delivers good optimization results, while maintaining a fast convergence speeds.

Dropout is a simple and yet effective regularization technique proposed in [23] to prevent a deep learning model from complex co-adaptations on training data, namely overfitting. It randomly ignores neurons to a predefined ratio (i.e., dropout rate) during training. During the forward propagation of each training step, the ignored neurons can not contribute to the activation of their connected neurons in latter layer temporally. Moreover, the weights of the ignored neurons is not updated during the back propagation. In our case, dropout has been applied to the last fully-connected layer in all implementations and a dropout rate of 0.5 is chosen. The choice of this hyperparameter is based on trial-and-error.

After the best validation accuracy, achieved in the 30$^{th}$ training epoch, both validation accuracy and loss, start to oscillate around the same level. This indicates that the model already fit the data to its best.

### B. Performance

The proposed CNN achieves the best performance and outperforms the boosting-based approach. Both training and test accuracies achieved accuracies of 98.34% and 99.65%, respectively. Table II, illustrates the confusion matrix of the trained CNN model performed on the test set. Accordingly, the misclassification in both classes is about 0.5% of the cases, which is suitable for the required task.

TABLE II: Human-Robot confusion matrix of CNN.

| True/Predicted | Human | Robot |
| --- | --- | --- |
| Human | 99.40% | 0.56% |
| Robot | 0.60% | 99.44% |

## VI. CONCLUSION

This work presents the use of FMCW radars to distinguish humans from robots in an industrial environment based on their velocity and range distributions. Since the required task involves aspects of human safety, a real-time detection technique is proposed based on R-D maps that consume much shorter time intervals than $\mu - D$ signatures. Accordingly, the use of R-D maps increases the number of datasets used for designing classifiers, thus a more robust classification is expected. Moreover, the range dimension can be used as an additional feature in classification and can help in determining how far a target is from the radar. Based on the target's classified type and range from the radar, special actions can be taken in such situations.

About 10,000 equally distributed R-D maps are collected from different human and industrial robot subjects. To generalize the classification on different surrounding clutters and noise, the measurements are all taken in different test areas. Furthermore, during each experiment test subjects are moved in front of the radar with different aspect angles to address different motion patterns and angles of incident. The collected R-D maps are then applied to different machine learning and deep learning approaches.

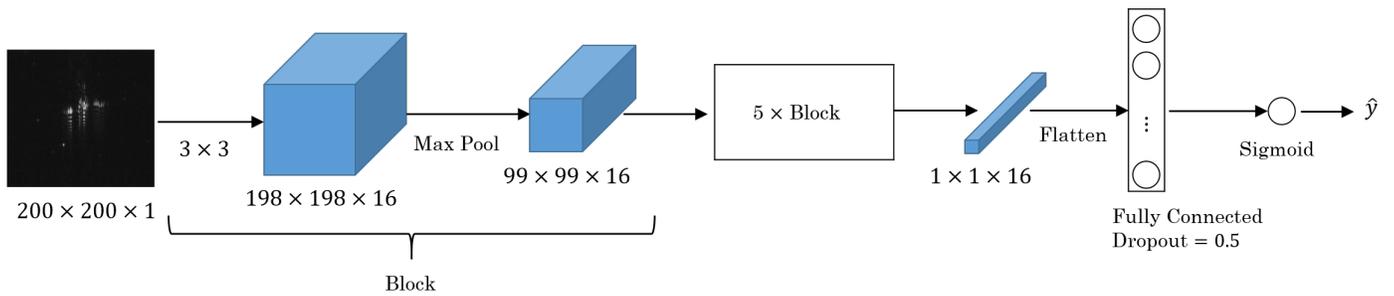

Fig. 6: Proposed CNN architecture on R-D maps.

Conventional learning techniques are first applied for the "Human/Robot" classification problem based on handcrafted features. The features are extracted to include the distribution in both R-D dimensions without specific ranges or velocities. Seven features are extracted from each R-D map and applied to classical learning approaches as SVM and K-NN. Unfortunately, the achieved performance at one R-D map is always lower than 90% which is not sufficient for our task. To achieve an acceptable performance, the seven features are extracted over successive R-D maps and concatenated as a sequence of feature vectors. In that case, SVM attained the best test accuracy of 95% on 10 successive buffers. Although the achieved accuracy is acceptable, 10 buffers requires a latency of 1 s on our current radar parametrization. This reflected that conventional techniques affects the required real-time aspect. This motivated the demonstration of a novel ensemble tree learning classifiers on restructured R-D data. In this part, a 2-dimensional mean is computed on each R-D map to get range and Doppler profiles. The extracted profiles are then shifted and restructured to include the patterns without exact velocity or range values. The restructured R-D data is concatenated in one vector as 512 samples and fed into random forest and gradient boosting classifiers. The gradient boosting attains a performance of 97% on one R-D map, outperforming the random forest and conventional techniques. Finally, a 6 layers CNN was trained on grayscale R-D maps and the trained network was shown to outperform all other techniques with a test accuracy of 99% on one R-D map (latency of 0.1 s).

This work addresses only the classification of a single moving target in the radar area. Extending the idea with Multiple-Input Multiple-Output (MIMO) radars can include estimating the angle of arrival; thus, the exact position of a moving target can be identified. Based on such positioning, the detection and classification of multiple subjects in the test area can be addressed. The presented approaches included the classification of moving human and robots only. However, static human detection must also be considered by means of vital signs or detected radar cross sections.